\documentclass{article}

\usepackage[final,nonatbib]{neurips_2019_ml4ps}

\usepackage[utf8]{inputenc} % allow utf-8 input
\usepackage[T1]{fontenc}    % use 8-bit T1 fonts
\usepackage{hyperref}       % hyperlinks
\usepackage{url}            % simple URL typesetting
\usepackage{booktabs}       % professional-quality tables
\usepackage{amsfonts}       % blackboard math symbols
\usepackage{nicefrac}       % compact symbols for 1/2, etc.
\usepackage{microtype}      % microtypography
\usepackage{amssymb}
\usepackage{xcolor}
\usepackage[ruled,vlined,linesnumbered]{algorithm2e}
\usepackage{empheq}
\title{Learning Generalized Quasi-Geostrophic Models Using Deep Neural Numerical Models}

\author{
  Redouane Lguensat, Julien Le Sommer, Sammy Metref, Emmanuel Cosme \\
  Universit\'e Grenoble Alpes, CNRS, IRD, Grenoble INP, IGE; 38000 Grenoble, France.\\
  \texttt{\{firstname.lastname\}@univ-grenoble-alpes.fr} \\
   \And
   Ronan Fablet \\
   IMT Atlantique; Lab-STICC, Brest, France \\
   \texttt{ronan.fablet@imt-atlantique.fr} \\
}
\begin{document}

\maketitle

\begin{abstract}
  %The abstract paragraph should be indented \nicefrac{1}{2}~inch (3~picas) on both the left- and right-hand margins. Use 10~point type, with a vertical spacing (leading) of 11~points.  The word \textbf{Abstract} must be centered, bold, and in point size 12. Two line spaces precede the abstract. The abstract must be limited to one paragraph.
We introduce a new strategy designed to help physicists discover hidden laws governing dynamical systems.
%Based on physical a priori knowledge and the power of numerical schemes, 
We propose to use machine learning automatic differentiation libraries to develop hybrid numerical models that combine components based on prior physical knowledge with components based on neural networks.
In these architectures, named Deep Neural Numerical Models (DNNMs), the neural network components are used as building-blocks then deployed for learning hidden variables of underlying physical laws governing dynamical systems. 
In this paper, we illustrate an application of DNNMs to upper ocean dynamics, more precisely the dynamics of a sea surface tracer, the Sea Surface Height (SSH).
We develop an advection-based fully differentiable numerical scheme, where parts of the computations can be replaced with learnable ConvNets, and make connections with the single-layer Quasi-Geostrophic (QG) model, a baseline theory in physical oceanography developed decades ago. 
\end{abstract}

\section{Introduction}

Physical modeling is still one of the most striking examples where humans are a long way ahead of pure Machine Learning (ML) systems. Recently, numerous research efforts have been directed into designing ML algorithms, especially deep neural networks, that can learn the basic laws of physics from data \cite{wu2018toward,iten2018discovering,greydanus2019hamiltonian,wang2019emergent}. These works mostly agree on the importance of interpretability and respect of physical constraints, which is still not straightforward when using black-box regressors such as neural networks \cite{carleo2019machine}. In particular, many works focused on the case where data describing the dynamical system of interest is assumed to be governed by a system of partial differential equations (PDEs) and where a certain physical a priori is known \cite{brunton2016discovering,raissi2018deep,long2019pde,lu2019extracting}. From the perspective of ocean sciences, many results shown in the aforementioned references were run on toy models, and more investigation is needed on real ocean satellite-derived or model simulation data. Few works set the foot in this direction: in \cite{de2017deep} ideas from optical flow video prediction were linked to an advection-diffusion model and considered to forecast Sea Surface Temperature (SST), while in \cite{ayed2019learning,ouala2019learning} the goal was to infer the dynamics of a latent variable from partial and noisy observations of SST and Sea Level Anomaly (SLA) respectively.

The general idea behind the present work consists in standing on the shoulders of the current understanding of ocean variables by physical oceanographers, and include as much as we can of their knowledge in the design of our NN architecture. In this paper, we demonstrate this strategy on upper ocean dynamics, and more precisely the dynamics of the Sea Surface Height (SSH). We present a fully-differential advection-diffusion architecture which generalizes the Quasi-Geostrophy (QG) theory, one of the main baselines for forecasting SSH \cite{ubelmann2015dynamic}. To ensure numerical stability and stay close to realistic solutions, ideas from numerical schemes were considered in developing the architecture at the expense of depth and memory load.

\section{Model}

\paragraph{Deep Neural Numerical Models}
Automatic differentiation (AD) has a long history in the numerical modeling community. In ocean sciences for instance, OpenAD an open source code for AD has been used to calculate adjoints of popular general circulation models such as MITgcm \cite{naumann2006adjoint,utke2008openad}. However, these tools do not account for the training of neural networks straightforwardly, and are in general designed to handle Fortran based codes. We refer by Deep Neural Numerical Models (DNNMs) to fully differentiable numerical models that can incorporate easily trainable NNs. Depending on the complexity of the involved PDE equations, corresponding DNNMs require a considerable amount of technical and engineering work, to rewrite standard numerical model codes into fully differentiable architectures that allows training NNs by backpropagation. This could not be possible without the ongoing flare in the Deep Learning community and in particular in AD librairies such as PyTorch and Tensorflow \cite{paszke2017automatic, abadi2016tensorflow,baydin2018automatic}. DNNMs can range from fully NN-based architectures such as ResNets \cite{weinan2017proposal,rousseau2019residual} to complex physical constrained architectures such as in \cite{raissi2018deep,de2017deep,long2019pde}. %For instance, the recent support of float64 precision was important since it is the standard number format in geophysical numerical models.
Here, we propose an advection-diffusion DNNM, that solves the following equations:

\begin{equation}
    a) \quad \Phi = \mathcal{T}_1(\Psi); \quad U = \mathcal{T}_2(\Psi); \quad V = \mathcal{T}_3(\Psi); \quad \quad b) \quad \frac{\partial \Phi}{\partial t} + U\frac{\partial \Phi}{\partial x} + V\frac{\partial \Phi}{\partial y} = D \nabla^2 \Phi
    \label{equ:psi},
\end{equation}

where $\nabla^2$ is the 2D Laplacian operator, $U$ and $V$ are components of the nondivergent velocity field, and  $D$ the diffusion coefficient. These equations describe the evolution of the flow field $\Psi$ through the advection-diffusion of a proxy variable $\Phi$ obtained by a given transformation $\mathcal{T}_1$. In case $\mathcal{T}_1$ is the identity, we fall into classical models as the one studied in \cite{de2017deep}. Using PyTorch, we develop a DNNM where the discretization of the PDE involves the use of a $3^{rd}$ order upwind scheme and a $1^{st}$ order Euler scheme in time. This scheme is stable as long as the Courant–Friedrichs–Lewy condition (CFL) is satisfied, implying that model integration is done in small steps $dt$. $\Psi$ at each $dt$ is obtained through the inversion of Equation \ref{equ:psi}a, for example if $\mathcal{T}_1$ is linear, we use Conjugate Gradient (CG) method with constant boundary conditions. In practice, given a good initial $\Psi$ guess, the CG is stopped after few iterations (less than 5) to permit real-time execution and to avoid computational burden. 

\begin{figure}[t]
    \centering
    \includegraphics[width=\linewidth]{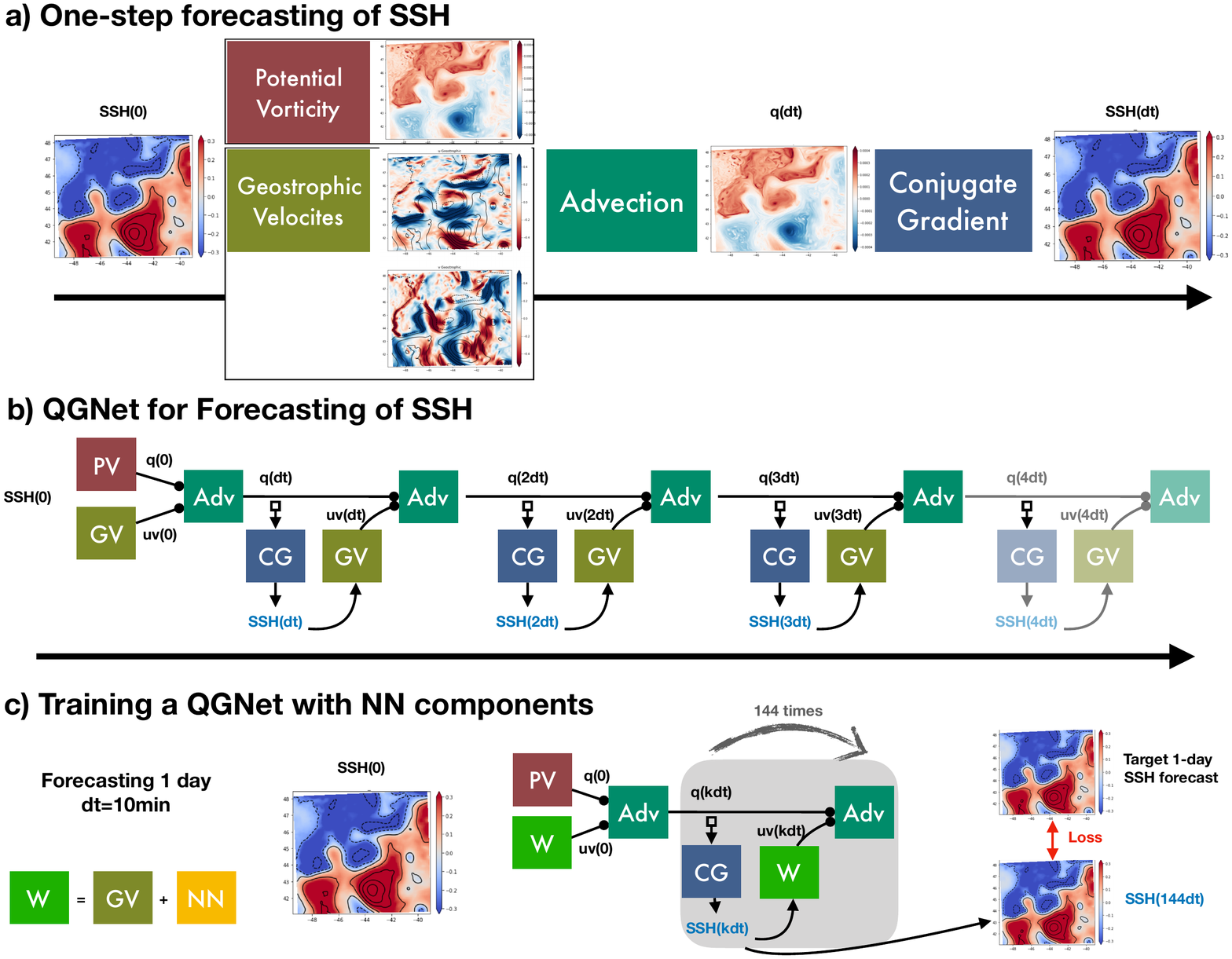}
    \caption{a) 1-step integration of QG equations; b,c) use cases of QG-Net}
    \label{fig:integ}
\end{figure}

\paragraph{QG-Net} %\textcolor{blue}{(((((((( add here two/three lines about QG)))))}
The 1-layer Quasi-geostrophic (QG) model is a reduced model that describes the evolution of oceanic flows close to geostrophic balance \cite{Vallis06}. Recently, this model was found to be a good baseline to dynamically interpolate SSH fields through temporal gaps \cite{ubelmann2015dynamic}. To forecast SSH dynamics, the numerical model uses $q$ the Potential Vorticity (PV) as a proxy variable that is advected by Geostrophic Velocities (GV). This process is governed by the following equations \cite{smith2001scales,hua1986numerical,fu1980nonlinear}: 

\begin{equation}
   \quad q = \frac{g}{f} (\nabla^2 h - \frac{h}{L^2_R}); \quad U^g =  \frac{-g}{f} \frac{\partial h}{\partial y}; \quad V^g = \frac{g}{f} \frac{\partial h}{\partial x}; \quad \frac{\partial q}{\partial t} + U^g\frac{\partial q}{\partial x} + V^g\frac{\partial q}{\partial y} + \beta V^g=0,
  \label{equ:PV}
\end{equation}

where $h$ is the SSH field, $g$ is the gravity constant, $f$ is the Coriolis parameter, $L_R$ is the first Rossby deformation radius. $U^g$ and $V^g$ are the Geostrophic Velocities, and $\beta V^g$ is a term that accounts for meridional advection of PV. This model is in accordance with the DNNM framework presented above. We find the correspondant $\mathcal{T}_1$, $\mathcal{T}_2$, $\mathcal{T}_3$ and call it QG-Net. Gradients and Laplacians were rewritten into filter convolutions. One CG iteration can already give an acceptable solution if the guess field respects the following pattern:
\begin{equation}
    h^{guess}_{dt}=h_0; \quad h^{guess}_{kdt} = 2 * h_{(k-1)dt} - h_{(k-2)dt} \quad k \in \mathbb{N}^* \backslash \{1\},
\end{equation}
therefore we used the equations of the CG algorithm \cite{shewchuk1994introduction} to write the equivalent of one CG iteration. The resulting architecture is illustrated in Fig\ref{fig:integ}(a,b). It benefits from high flexibility, since several building blocks can be interchanged with ConvNets, making it a playground for several modeling choices such as the ones exposed in the next section. Yet, due to the high memory cost of using float64 precision and low values of $dt$ needed for the integration step (which results in a high number of iterations), training QG-Net in an acceptable time requires several high-end Graphics Processing Units (GPUs).

%The overall algorithm is described in Algorithm \ref{alg:QG}. %The code can be found in the following Github repository \url{https://github.com/redouanelg/qgsw-DI}.

\section{Experiments}

\paragraph{Data and Experimental details}
%\textcolor{blue}{(((((help describe NATL60))))}
We use NATL60, a dataset from a comprehensive realistic ocean model simulation based on NEMO ocean engine run at kilometric resolution over the North Atlantic basin \cite{jean_marc_molines_2018_1210116}.
Study region is a $10^\circ \times 10^\circ$ box located on the Gulf Stream, a region with challenging physics. Four Nvidia Tesla V100 GPUs were used for the computations described in the experiments.

\paragraph{Discovering insights about hidden laws from data} NATL60 ocean circulation model is governed by complex physics not covered entirely by the QG theory. Yet, we want to investigate to which extent QG-Net can reveal the limits of this theory. A simple illustrative example consists in assuming that the PV is advected by some unknown fields $U^*$ and $V^*$ which are first-order derivations of SSH through a linear operator $F$. Concretely, given that the 2D gradient filter used to calculate the geostrophic velocities $V^g$ and $U^g$ in the original Python code are respectively $F_{QG} = \begin{pmatrix} 
-0.25 & 0 & 0.25 \\
-0.25 & 0 & 0.25
\end{pmatrix}$ and $F_{QG}^T$ its transpose, we replace $F_{QG}$ in QG-Net by a 6-parameters trainable filter and retrieve the resulting filter from a training procedure using NATL60 data. This resorts to:
\begin{equation}
    U^* = - \frac{g}{f} \frac{F^T \circledast SSH}{\partial y} \quad V^* = \frac{g}{f} \frac{F \circledast SSH}{\partial x}
\end{equation}
We set a 1-day SSH forecasting experiment, and use $dt=10min$ for QG-Net, meaning that 144 blocks are needed (Fig1 (b,c)). Note that we use shared weights across the blocks. QG-Net in this experiment has then only 6 parameters which are the weights of $F$. 18 SSH images of size $200\times 150$ are used for training (one each 20 days from June 2012 to June 2013) using the BFGS algorithm and a loss function $\mathcal{L}$ composed of three terms: i) mean square error between the QG-Net 1-day forecast and the NATL60 target 1-day forecast scaled by the variance of the target, ii) a loss penalizing velocity fields with high divergence, iii) L2-regularization of the weights
\begin{equation}
\mathcal{L} = \frac{1}{n}\Sigma_{i=1}^{n}{\Big(\frac{SSH^{1day}_{i,NATL60} -SSH^{1day}_{i,QGNet}}{\sigma_i(SSH^{1day}_{NATL60})}\Big)^2} + \lVert \nabla . ([U_0^*,V_0^*]) \lVert_{2}^2 + 10^{-3} \lVert F \lVert_{2}^2
\label{loss}
\end{equation}

The result of the optimization yields $F = \begin{pmatrix} 
-0.2629 & 0.0029 & 0.2592 \\
-0.2099 & -0.0008 & 0.2124
\end{pmatrix}$, keep in mind that we are not expecting to find exactly $F_{QG}$ due to the complex dynamics of NATL60. Therefore, from a completely random filter, QG-Net found $F$ which is close to $F_{QG}$, a proof perhaps that the capacity of this type of models is reached and that the PV is best advected by the GV, as QG-theory claims.

\begin{figure}[t]
    \centering
    \includegraphics[width=\linewidth]{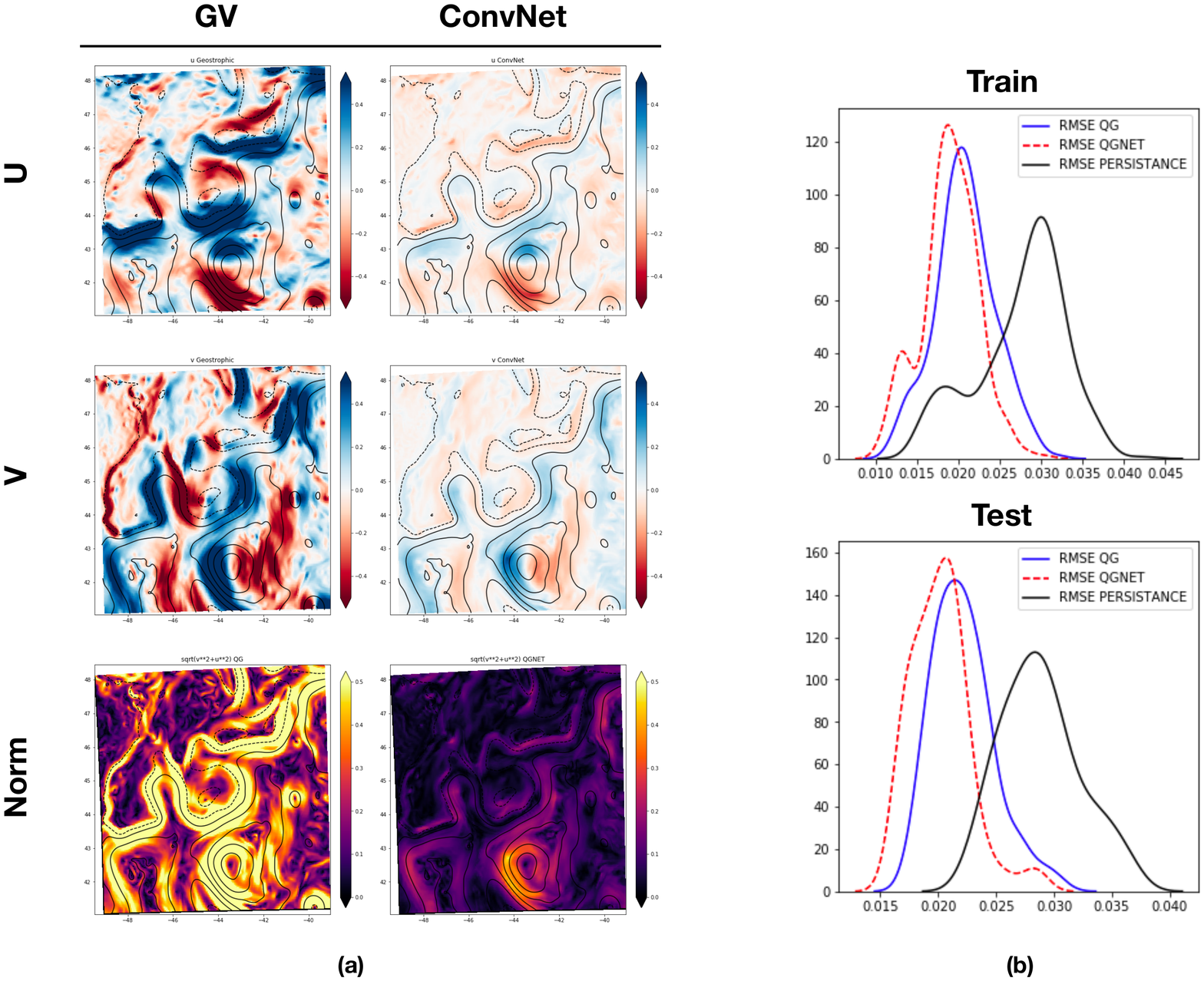}
    \caption{a) An example of the Geostrophic velocities, their norm and the nonlinear additive components produced by the trained ConvNet. SSH contours are shown in black. b) RMSE distributions on the train and test datasets}
    \label{fig:exp2}
\end{figure}

\paragraph{Supplementing known physics with nonlinear learnable components} In this experiment, we consider the same 1-day SSH forecast experiment but we assume that at each time step the PV is advected by the GV plus NN a nonlinear transformation of SSH (Fig1(c)). NN is a 2-layer ConvNet with 16 $\{3\times3\}$ filters, Batch Normalization and leaky ReLU activations, the output layer is a linear layer with 2 channels that are added to $U^g$ and $V^g$ respectively to form W the new velocity fields. QG-Net has 2545 trainable parameters and we split our data into 122 SSH images of size $200\times 150$ (1 SSH map each 3 days from 14 Jun 2012 to 13 Jun 2013), then after a 10-day gap we take 32 snapshots as our test dataset (1 SSH map each 3 days from 24 Jun to 29 Sep 2013). Our network is trained using Adam optimizer with an initial learning rate of $1e^{-3}$ which is later multiplied by 0.1 each 100 epoch. Batch size is 4 samples distributed on the 4 GPUs cards. The loss function considered here is the scaled mean square error used in Eq.\ref{loss}. To ensure a stable gradient flow at the beginning of the optimization a scalar parameter initialized as zero is multiplied to the ConvNet velocities.

At the end of training, we unplug the ConvNet from QG-Net, resulting in a NN component that takes SSH as input and yields a deterministic "perturbation" of GV. A clear benefit from this setup is that the trained component can be plugged back in the original Python code and avoid computational load at test time. Fig\ref{fig:exp2}(b) presents the RMSE distributions of the standard QG, our QG-Net and a naive constant model (persistance). Adding the ConvNet component to the GV has slightly improved the standard QG model, this is an indication that a nonlinear velocity term can model SSH dynamics beyond the standard QG. Fig\ref{fig:exp2}(a) shows an example of the additional velocities produced by the ConvNet along with GV for the same SSH input. We observe that the output of the ConvNet has a significant amplitude along SSH contours (ocean fronts), and that the fields follow a special pattern that depends on GV and are not completely random. Interpreting the ConvNet in this experiment is not straightforward and calls for more investigation to convert it into tangible equations that could be inspected by physical oceanographers. 

%\subsection{Figures}

%\begin{figure}
%  \centering
%  \fbox{\rule[-.5cm]{0cm}{4cm} \rule[-.5cm]{4cm}{0cm}}
%  \caption{Sample figure caption.}
%\end{figure}

%All artwork must be neat, clean, and legible. Lines should be dark enough for purposes of reproduction. The figure number and caption always appear after the figure. Place one line space before the figure caption and one line space after the figure. The figure caption should be lower case (except for first word and proper nouns); figures are numbered consecutively.

%You may use color figures.  However, it is best for the figure captions and the paper body to be legible if the paper is printed in either black/white or in color.

%\section{Related Work}

\section{Conclusion}
%This work started from the following questions: \textbf{can already established physical theories be rediscovered using only data and machine learning algorithms? can we do even better?}. 
We show that combining deep learning automatic differentiation libraries and numerical models could help designing hybrid models with trainable parameters and represent a test bed to evaluate established physical theories or seek intuition for developing new ones. 
%While it is still very early to talk about AI systems than can replace human scientists. 
We believe that this work represents a modest step for helping physicists developing innovative physical models. 

\subsubsection*{Acknowledgments}
The authors would like to thank Clément Ubelmann from CLS for the 1-layer QG Python code that can be found here \url{https://github.com/redouanelg/qgsw-DI}. Most of the computations presented in this paper were performed using the GRICAD infrastructure (\url{https://gricad.univ-grenoble-alpes.fr}), which is partly supported by the Equip@Meso project (reference ANR-10-EQPX-29-01) of the programme Investissements d'Avenir supervised by the Agence Nationale pour la Recherche. \\R. Lguensat is funded through a postdoctoral grant from Centre National d'Etudes Spatiales (CNES), he also acknowledges the support of NVIDIA Corporation under the NVIDIA GPU Grant program.
S. Metref is funded by ANR through contract number ANR-17- CE01-0009-01.
R. Fablet was supported by Labex Cominlabs (grant SEACS), CNES (grant OSTST-MANATEE) and ANR (EUR Isblue and Melody).

%Use unnumbered third level headings for the acknowledgments. All acknowledgments go at the end of the paper. Do not include acknowledgments in the anonymized submission, only in the final paper.

%\section*{References}
%\newpage
\small
\bibliographystyle{unsrt}
\bibliography{neurips_2019}

\end{document}